%% file: 000_cvpr.tex
\documentclass[final]{cvpr}

\usepackage{times}
\usepackage{epsfig}
\usepackage{graphicx}
\usepackage{amsmath}
\usepackage{amssymb}
\usepackage{caption}

\usepackage{multirow}
\usepackage{enumerate}
\usepackage{color}
\usepackage{xspace}
\usepackage{booktabs}
\usepackage{makecell}

\newcommand{\armthor}{{ManipulaTHOR}\xspace}
\newcommand{\task}{\textsc{ArmPointNav}\xspace}
\newcommand{\dataset}{\textsc{APND}\xspace}
\newcommand{\pnav}{\textsc{PointNav}\xspace}
\newcommand{\thor}{\textsc{AI2-THOR}\xspace}

\usepackage[pagebackref=true,breaklinks=true,colorlinks,bookmarks=false]{hyperref}

\def\cvprPaperID{3887} 
\def\confYear{CVPR 2021}

\providecommand{\spacebet}{\hspace{8pt}}

\begin{document}

\title{\armthor: A Framework for Visual Object Manipulation}

\author{Kiana Ehsani$^{1,2}$ \spacebet
Winson Han$^1$\spacebet
Alvaro Herrasti$^1$\spacebet
Eli VanderBilt$^1$\spacebet
Luca Weihs$^1$\\
Eric Kolve$^1$\spacebet
Aniruddha Kembhavi$^{1,2}$\spacebet
Roozbeh Mottaghi$^{1,2}$\\
$^1$ Allen Institute for AI\spacebet \spacebet $^2$ University of Washington\\
\url{https://ai2thor.allenai.org/manipulathor}
}

\twocolumn[{
\renewcommand\twocolumn[1][]{#1}
\maketitle
\vspace*{-0.5cm}
\centering
\includegraphics[width=.99\linewidth]{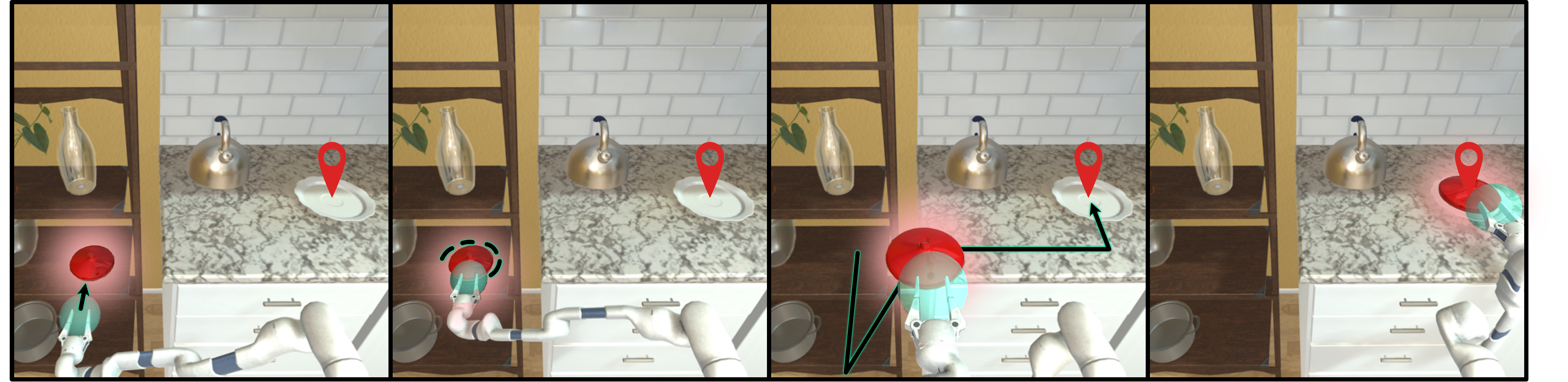}
\vspace{-.2cm}
\captionof{figure}{We address the problem of \emph{visual object manipulation}, where the goal is to move an object between two locations in a scene. Operating in visually rich and complex environments, generalizing to unseen environments and objects, avoiding collisions with objects and structures in the scene, and visual planning to reach the destination are among the major challenges of this task. Here, we illustrate a sequence of actions taken a by a virtual robot within the AI2-THOR environment for picking up a vase from the shelf and stack it on a plate on the countertop.}
\label{fig:teaser}
\vspace*{0.3cm}
}] 

\maketitle
\thispagestyle{empty}

\begin{abstract}

   \input{text/00_abstract}
\end{abstract}


\input{text/01_intro}
\input{text/02_related}
\input{text/03_thor}

\input{text/04_dataset}
\input{text/05_method}

\input{text/06_experiments}
\input{text/07_discussion}

\noindent \small{\textbf{Acknowledgement.} We thank Dustin Schwenk for suggesting the name of the framework.}

{\small
\bibliographystyle{ieee_fullname}
\bibliography{egbib}
}

\input{text/08_supplementary}

\end{document}

%% file: text/00_abstract.tex
\vspace{-1em}
The domain of Embodied AI has recently witnessed substantial progress, particularly in navigating agents within their environments. These early successes have laid the building blocks for the community to tackle tasks that require agents to actively interact with objects in their environment. Object manipulation is an established research domain within the robotics community and poses several challenges including manipulator motion, grasping and long-horizon planning, particularly when dealing with oft-overlooked practical setups involving visually rich and complex scenes, manipulation using mobile agents (as opposed to tabletop manipulation), and generalization to unseen environments and objects. We propose a framework for object manipulation built upon the physics-enabled, visually rich AI2-THOR framework and present a new challenge to the Embodied AI community known as ArmPointNav. This task extends the popular point navigation task~\cite{anderson18} to object manipulation and offers new challenges including 3D obstacle avoidance, manipulating objects in the presence of occlusion, and multi-object manipulation that necessitates long term planning. Popular learning paradigms that are successful on PointNav challenges show promise, but leave a large room for improvement.

%% file: text/01_intro.tex
\section{Introduction}

Embodied AI, the sub-specialty of artificial intelligence at the intersection of robotics, computer vision, and natural language processing continues to gain popularity amongst researchers within these communities. This has expedited progress on several fronts -- open source \emph{simulators} are getting faster, more robust, and more realistic via photorealism and sophisticated physics engines, a variety of \emph{tasks} are being worked on such as navigation and instruction following, new \emph{algorithms} and \emph{models} are inching us towards more powerful and generalizable models and the recent development of multiple sim-to-real environments with paired worlds in simulation and real is enabling researchers to study the challenges of overcoming the domain gap from virtual to physical spaces. A notable outcome has been the development of near-perfect pure learning-based Point Navigation \cite{wijmans2019dd} agents, far outperforming classical approaches.

Most of the focus and progress in Embodied AI has revolved around the task of navigation -- including navigating to coordinates, to object instances, and to rooms. Navigating around in an environment is a critical means to an end, not an end in itself. The aspiration of the Embodied AI community remains the development of embodied agents that can perform complex tasks in the real world, tasks that involve actively manipulating objects in one's environment. The early successes and interest in Embodied AI have laid a foundation for the community to tackle the myriad of challenges that lie within the problem of object manipulation.

Object manipulation has long posed daunting challenges to roboticists. Moving manipulators within an environment requires estimating free spaces and avoiding obstacles in the scene, tasks which are rendered even harder due to the unwieldy nature of robotic arms. Generalizing to novel environments and objects is another important challenge. Finally, real-world tasks often involve manipulating multiple objects in succession in cluttered scenes, which requires fairly complex visual reasoning and planning. Besides, developing simulators for object manipulation poses a unique set of challenges. In contrast to navigation tasks that require camera translation and fairly rudimentary collision checking, object manipulation requires fine grained collision detection between the agent, its arms, and surrounding objects, and the usage of advanced physics emulators to compute the resulting displacements of the constituent entities. In particular, these computations are expensive and require significant engineering efforts to produce effective simulations at reasonably high frame rates.

We extend the AI2-THOR \cite{kolve2017ai2} framework by adding arms to its agents, enabling these agents to not only navigate around their environments but also actively manipulate objects within them. The newly introduced arm rig is designed to work with both forward and inverse kinematics, which allows one to control the arm using both joint actuations or by specifying the desired wrist translation. This flexibility allows Embodied AI practitioners to train policies requiring fine-grained actuator controls for all joints if they so desire, or instead use inbuilt kinematics functionalities and focus solely on the desired positioning of the end of the arm and manipulator. 

As a first step towards generalizable object manipulation, we present the task of \task -- moving in the scene towards an objects, picking it up and moving it to the desired location (Figure~\ref{fig:teaser}). \task builds upon the navigation task of PointNav \cite{anderson18} in that it is an atomic locomotive task, a key component of more complex downstream goals, specifies source and target locations using relative coordinates as opposed to other means such as language or images and utilizes compass as part of its sensor suite. But in contrast, it offers significant new challenges. Firstly, the task requires the motion of both the agent and the arm within the environment. Secondly, it frequently entails reaching behind occluding obstacles to pick up objects which requires careful arm manipulation to avoid collisions with occluding objects and surfaces. Thirdly, it may also require the agent to manipulate multiple objects in the scene as part of a successful episode, to remove objects, or make space to move the target object, which requires long-term planning with multiple entities. Finally, the motion of the arm frequently occludes a significant portion of the view, as one may expect, which is in sharp contrast to PointNav that only encounters static unobstructed views of the world.

The end-to-end \task model provides strong baseline results and shows an ability to not just generalize to new environments but also to novel objects within these environments -- a strong foundation towards learning generalizable object manipulation models. This end-to-end model is superior to a disjoint model that learns a separate policy for each skill within an episode.

In summary, we (a) introduce a novel efficient framework (\armthor) for low level object manipulation, (b) present a new dataset for this task with new challenges for the community, and (c) train an agent that generalizes to manipulating novel objects in unseen environments. Our framework, dataset and code will be publicly released. We hope that this new framework encourages the Embodied AI community towards solving complex but exciting challenges in visual object manipulation.

%% file: text/02_related.tex
\section{Related Works}
\noindent \textbf{Object Manipulation.} A long-standing problem in robotics research is object manipulation \cite{fearing1986implementing,bicchi2000robotic,bohg2013data,Stilman2007ManipulationPA,yahya2017collective,bousmalis2018using,fang2018multi,levine2018learning,fang2020learning}. Here, we explain some recent example works that are more relevant to our work. \cite{garrett2020online} address the problem of multi-step manipulation to interact with objects in presence of clutter and occlusion. \cite{murali20206} propose a planning approach to grasp objects in a cluttered scene by relying on partial point cloud observation. \cite{xu2020learning} learn a 3D scene representation to predict the dynamics of objects during manipulation. \cite{Haarnoja2018ComposableDR} propose a reinforcement learning approach for robotic manipulation where they construct new policies by composing existing skills. \cite{fang20} propose a model-based planner for multi-step manipulation. \cite{Li2019HRL4INHR,Xia2020ReLMoGenLM} study mobile manipulation by generating sub-goal tasks.
A combination of visually complex scenes, generalization to novel objects and scenes, joint navigation and manipulation, as well as navigating while manipulating object in hand are the key factors that distinguish our work from the previous work on object manipulation.

\noindent \textbf{Environments for object manipulation.} 
While several popular Embodied AI frameworks have focused on the navigation task, recently proposed improvements and frameworks such as iGibson \cite{xia2020interactive}, SAPIEN \cite{xiang2020sapien} and TDW \cite{gan2020threedworld} have enabled new research into manipulation.
Sapien \cite{xiang2020sapien} is a virtual environment designed for low-level control of a robotic agent with an arm. In contrast, our framework includes a variety of visually rich and reconfigurable scenes allowing for a better exploration of the perception problem. Meta-World \cite{yu2020meta} is developed to study multi-task learning in the context of robotic manipulation. The Meta-World framework includes a static table-top robotic arm and a limited set of objects. In contrast, our framework enables studying the problem of joint navigation and manipulation using a variety of objects. RLBench \cite{james2020rlbench} also provides a simulated environment for a table-top robotic arm. RoboTurk \cite{mandlekar2018roboturk} is a crowdsourcing platform to obtain human trajectories for robotic manipulation. RoboTurk also considers table-top manipulation scenarios. \cite{gupta2018robot} provide a large-scale dataset of grasping and manipulation to evaluate the generalization of the models to unstructured visual environments. Unlike our framework, their dataset is non-interactive and includes only pre-recorded manipulation trajectories. iGibson \cite{xia2020interactive} involves object interaction, but it does not support low-level manipulation (the interactions primarily involved pushing objects and rotation around hinges). Recently, an extension of iGibson \cite{shen2020igibson} has enabled object manipulation with contact forces.

\noindent \textbf{Visual navigation.} Our problem can be considered as an extension of the visual navigation work \cite{zhu17,gupta17,mirowski17,savinov18,wortsman19,yang19,chaplot20,wijmans2019dd} in the Embodied AI literature. There are a few key differences between our manipulation task and navigation. In manipulation, the shape of the agent changes dynamically due to the extension of the arm. Also, the manipulation of objects is performed in 3D and through clutter, while the navigation works assume 2D motion on a plane in fairly clean scenes. Finally, our proposed task requires the agent to plan its motion as well as the motion of its arm simultaneously.

%% file: text/03_thor.tex
\section{\armthor}

The growing popularity of Embodied AI can be partly attributed to the availability of numerous free and fast 3D simulators such as \thor \cite{kolve2017ai2}, Habitat \cite{savva2019habitat} and iGibson \cite{xia2020interactive}. Some of these simulators excel at their photorealism, some at their speed, some at the interactivity they afford while others at their physics simulations. While researchers have many options to choose from when it comes to researching embodied navigation, fewer choices exist to study object manipulation, particularly in visually rich environments with varied objects and scenes. Simulating object manipulation presents unique challenges to simulator builders beyond ones posed by navigation, including the need for fine-grained physics emulations, object and manipulator properties, and obtaining acceptable frame rates.

We present \armthor, an extension to the \thor framework that adds arms to its agents. \thor is a suitable base framework due to its powerful physics engine, Unity, variety of realistic indoor scenes, large asset library of open source manipulable objects as well as articulated receptacles such as cabinets, microwaves, boxes, and fridges. While \thor has been previously used to train agents that interact with objects, this interaction has been invoked at a high level -- for instance, a cabinet is opened by choosing a point on the cabinet and invoking the ``open" command. \armthor allows agents to interact with objects at a lower level via their arm manipulators, and thus opens up a whole new direction for Embodied AI research.
The sensors that are available for use are RGB image, depth frame, GPS, agent's location, and arm configuration.

\begin{figure}[tp]
\centering
\includegraphics[width=19pc]{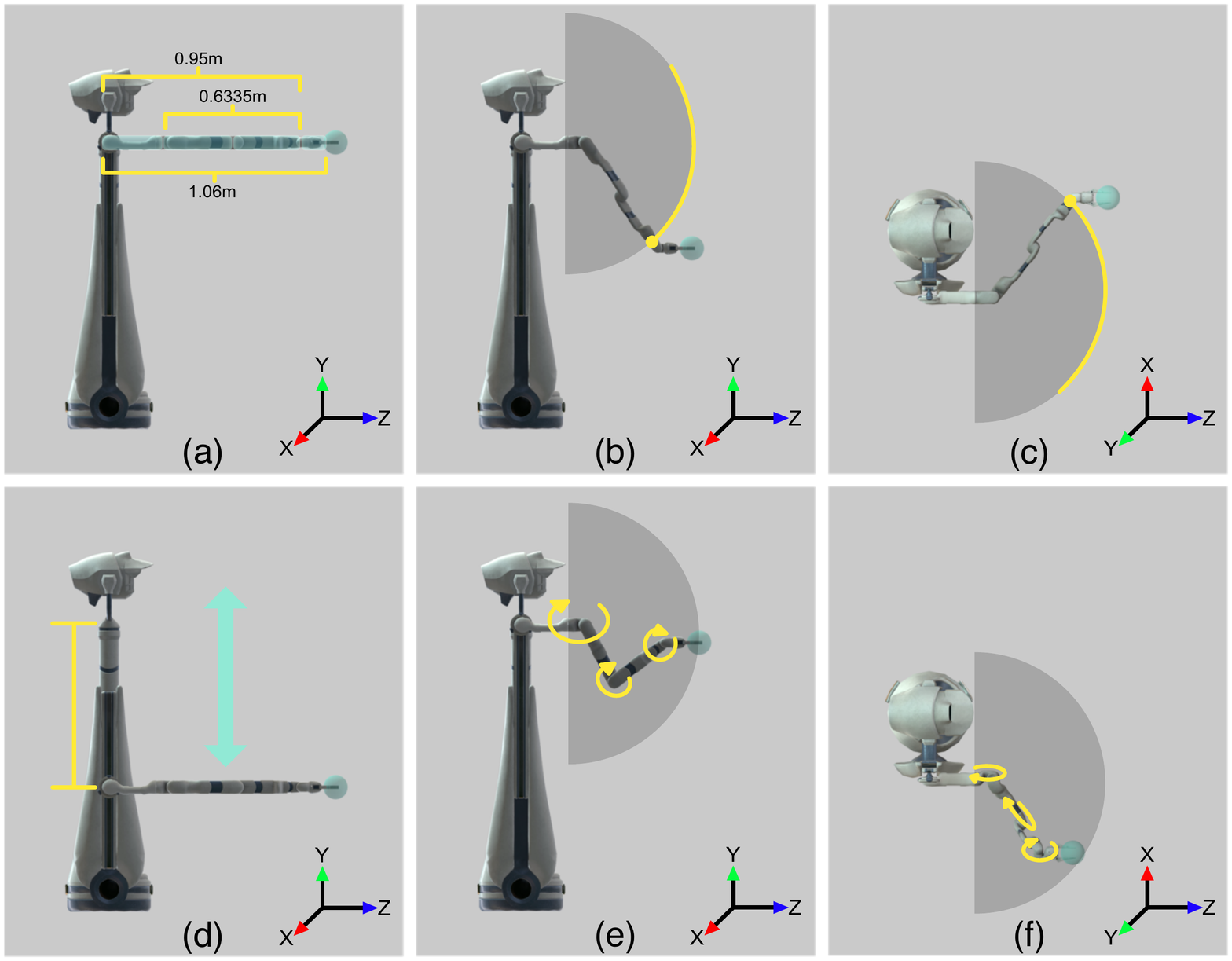}
\vspace{-1em}
\caption{\textbf{Arm Design and Kinematic Constraints.} The arm consists of four joints (a). The max reach of the arm is defined by a hemisphere centered around the end of the first joint, whose radius is equal to 0.6335m (b,c). The height of the arm can be adjusted along the body of the agent (d). All joint rotations are solved by inverse kinematics, so any position the wrist can move to within the hemisphere's extents will have the joints rotate to accommodate the position of the wrist joint (e,f).}
\vspace{-1em}
\label{fig:arm_constraint}
\end{figure}

\textbf{Arm Design.} In \armthor, each agent has a single arm. The physical design of the arm is deliberately simple: a three-jointed arm with equal limb-lengths, attached to the body of the agent. This design is inspired by Kinova’s line of robots~\cite{kinova}, with smooth contours and seamless joint-transitions and it is composed entirely of swivel joints, each with a single axis of rotation. The shoulder and wrist support 360 degrees of rotation and the hand grasper comes with a 6DOF (see Figure~\ref{fig:arm_constraint}). The robot's arm rig has been designed to work with either forward or inverse kinematics (IK), meaning its motion can be driven joint-by-joint, or directly from the wrist, respectively. 

\textbf{Grasper.} The Grasper is defined as a sphere at the end of the arm. Objects that intersect with the sphere can be picked up by the grasper. This abstract design follows the `abstracted grasping' actuation model of~\cite{batra2020rearrangement} in lieu of more involved designs like jaw grippers or humanoid hand graspers, enabling researchers to solve problems involved with object manipulation through the environment without having to account for the complexities of grasping. Object grasping is a challenging problem with a rich history in the robotics community and we hope to add this explicit functionality into \armthor in future versions. 

\textbf{Arm Interface.} The arm comes with the following functionalities: 1) manipulating the location and orientation of the wrist (the joints connecting the base of the arm to the wrist are resolved via IK as the wrist moves), 2) adjusting the height of the arm, 3) obtaining the arm state's metadata including joint positions, 4) picking up the objects colliding with the grasper's sphere, 5) dropping the held object and 6) changing the radius of the grasper's sphere.

\textbf{Physics Engine.} We use NVIDIA's PhysX engine through Unity’s engine integration to enable physically realistic object manipulation. This engine allows us to realistically move objects around, move the arm in the space, and cascade forces when the arm hits an object.

\textbf{Rendering Speed.} Accurate collision detection and object displacement estimation are very time consuming but are important requirements for our simulator. Through extensive engineering efforts, we are able to obtain a training speed of 300 frames per second (fps) on a machine with 8 NVIDIA T4 GPUs running 40 cores. To put this into perspective, \pnav using \thor on the same machine achieves a training speed of roughly 800 fps, but has very rudimentary collision checks and no arm to manipulate. At 300 fps researchers may train for 20M steps per day, a fast rate to advance research in this direction, which we hope to improve significantly with more optimizations in our code base.

%% file: text/04_dataset.tex
\section{\task}
\label{sec:dataset}

As a first step towards generalizable object manipulation, we present the task of \task -- moving an object in the scene from a source location to a target location. This involves, navigating towards the object, moving the arm gripper close to the object, picking it up, navigating towards the target location, moving the arm gripper (with the object in place) close to the target location, and finally releasing the object so it lands carefully. In line with the agent navigation task of \pnav \cite{anderson18}, source and target locations of the object are specified via $(x,y,z)$ coordinates in the agent coordinate frame.

\begin{figure}[h]
\centering
\includegraphics[width=20pc]{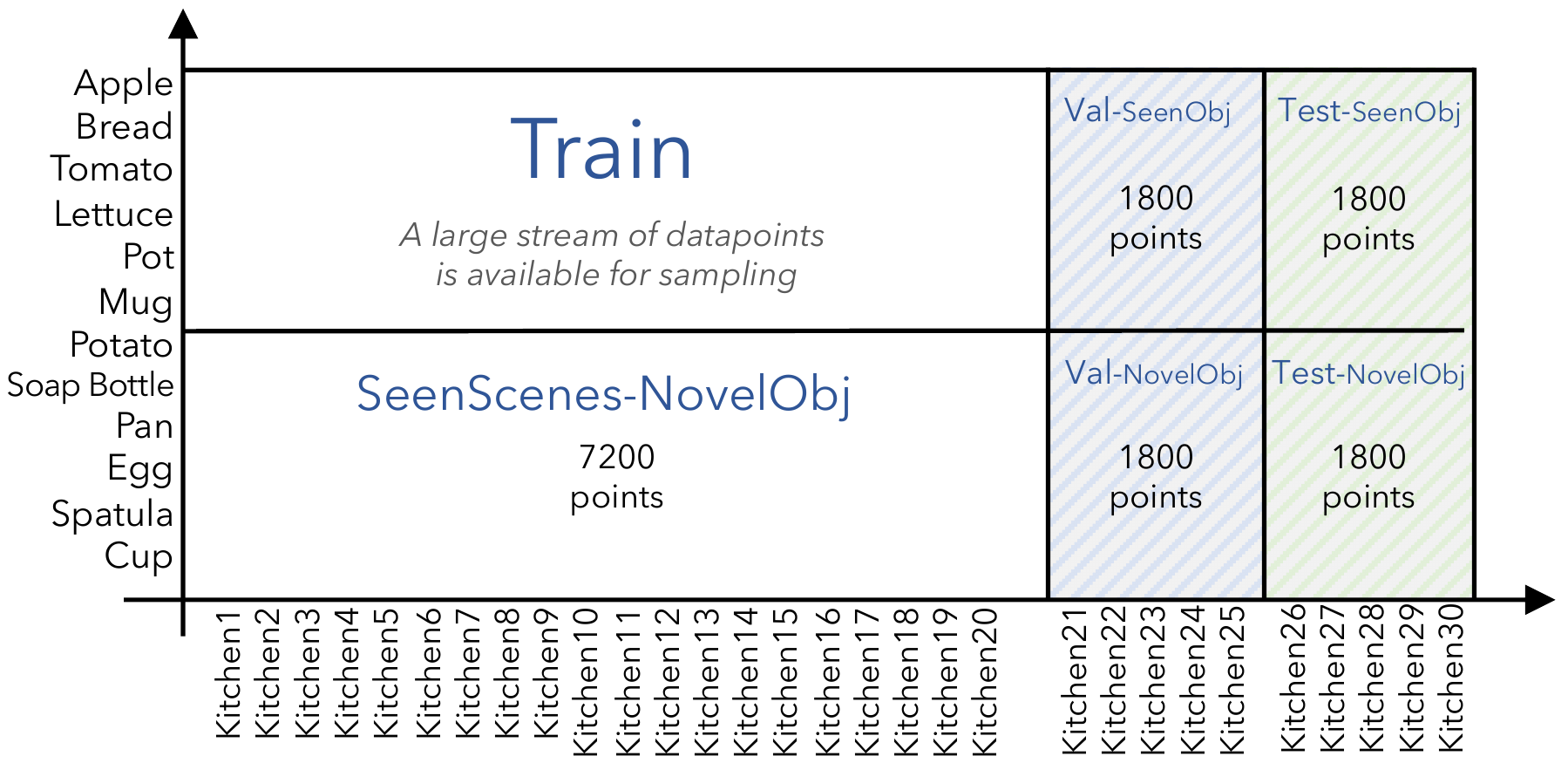}
\vspace{-1em}
\caption{\textbf{Scene and object splits in \dataset.} In order to benchmark the performance on \task, in addition to providing a large pool of datapoints for training, we provide a small subset of tasks per data split. We randomly subsampled 60 tasks per object per scene for evaluation purposes.
}
\vspace{-1em}
\label{fig:data_splits}
\end{figure}

\begin{figure*}[tp]
\centering
\includegraphics[width=30pc]{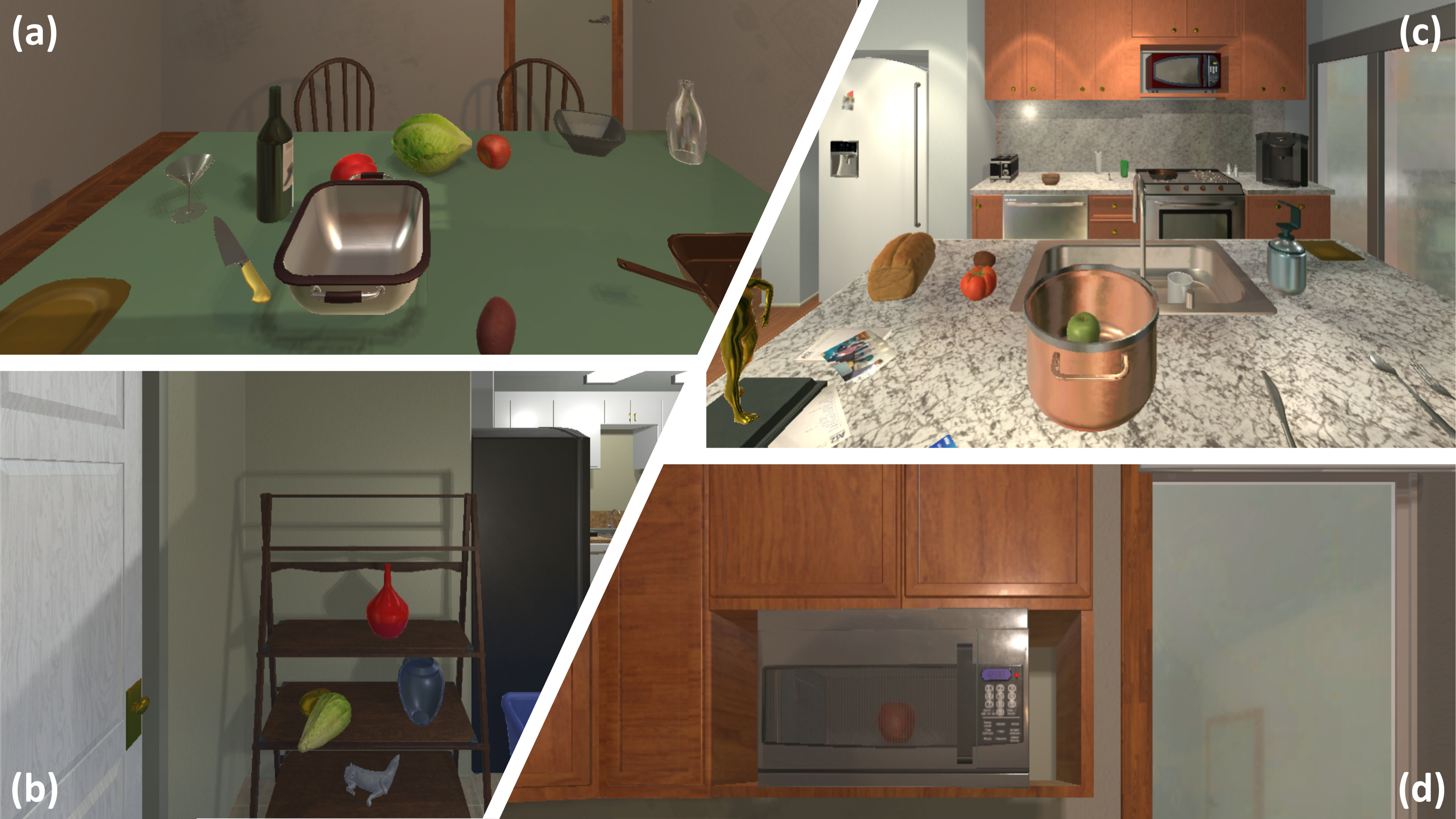}
\vspace{-1em}
\caption{\textbf{Dataset Samples.} The initial location of the object can pose a variety of challenges. In (a) the tomato is occluded by the bowl, therefore the agent needs to remove the bowl to reach the tomato. In (b) the lettuce is on the shelf, which requires the agent to move the arm carefully such that it does not collide with the shelf or the vase. In (c) The goal object is inside another object and in (d) the goal object is inside a receptacle, therefore it requires interacting with another entity (opening microwave's door) before reaching for the object. The latter case is outside the scope of this paper.}
\vspace{-1em}
\label{fig:dataset_sample}
\end{figure*}

\noindent \textbf{Dataset.} To study the task of \task, we present the Arm \pnav Dataset (\dataset). This consists of 30 kitchen scenes in \thor that include more than 150 object categories (69 interactable object categories) with a variety of shapes, sizes and textures. We use 12 pickupable categories as our target objects. As shown in Figure~\ref{fig:data_splits}, we use 20 scenes in the training set and the remaining is evenly split into Val and Test. We train with 6 object categories and use the remaining to test our model in a Novel-Obj setting.

\noindent \textbf{Metrics.} We report the following metrics: 
\\$\bullet$ Success rate without disturbance \textbf{(SRwD)} -- Fraction of successful episodes in which the arm (or the agent) does not collide with/move other objects in the scene.
\\$\bullet$ Success rate \textbf{(SR)} -- Similar to SRwD, but less strict since it does not penalize collisions and movements of objects. 
\\$\bullet$ Pick up success rate \textbf{(PuSR)} -- Fraction of episodes where the agent successfully picks up the object.
\\$\bullet$ Episode Length \textbf{(Len)} -- Episode length for both success and failure episodes.
\\$\bullet$ Successful episode Length \textbf{(SuLen)} -- Episode length for successful episodes.
\\$\bullet$ Pick up successful episode length \textbf{(PuLen)} -- Episode length for episodes with successful pickups.

\dataset offers significant new challenges. The agent must learn to navigate not only itself but also its arm relative to its body. Also, as the agent navigates in the environment, it should avoid colliding with other objects -- which brings new complexities given the addition of the arm and potentially carrying an object in its gripper. Further, reaching to pick up objects involves free-space estimation and obstacle avoidance -- which becomes challenging when the source or target locations are behind the occluders. Moreover, it needs to choose the perfect time to attempt pickup as well as ending the episode. Finally, these occluders themselves may need to be manipulated in order to complete the task. 
The agent should overcome these challenges while its view is frequently obstructed by the arm and/or the object being carried. Figure~\ref{fig:dataset_sample} illustrates a few of the challenges involved with picking up the object from its source location.

\begin{figure}[tp]
\centering
\includegraphics[width=20pc]{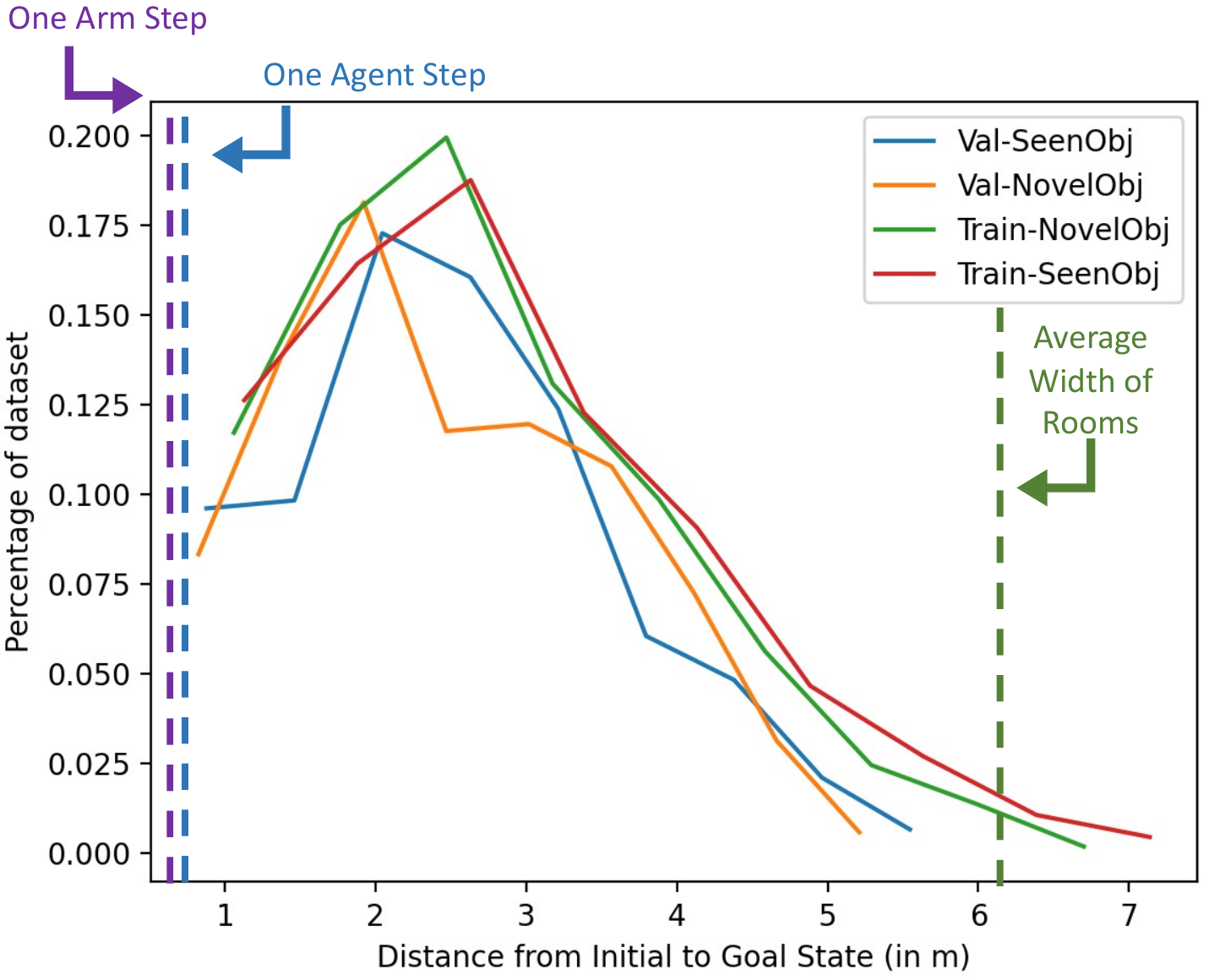} 
\vspace{-1em}
\caption{\textbf{Statistics of the dataset.} This plot presents the distribution of the initial distance of the object from the target location in meters. We mark the step size for the agent and the arm movements, and the average width of the rooms for reference.}
\vspace{-1em}
\label{fig:dataset_distances}
\end{figure}

Figure~\ref{fig:dataset_distances} shows the distribution of the distances of the target location of the object from its initial state. For comparison, we show the step size for agent navigation and arm navigation as well. Note that the initial distance of the agent from the object is not taken into account. 

%% file: text/05_method.tex
\section{Model}
\label{sec:method}

\begin{figure*}[tp]
    \centering
    \includegraphics[width=35pc]{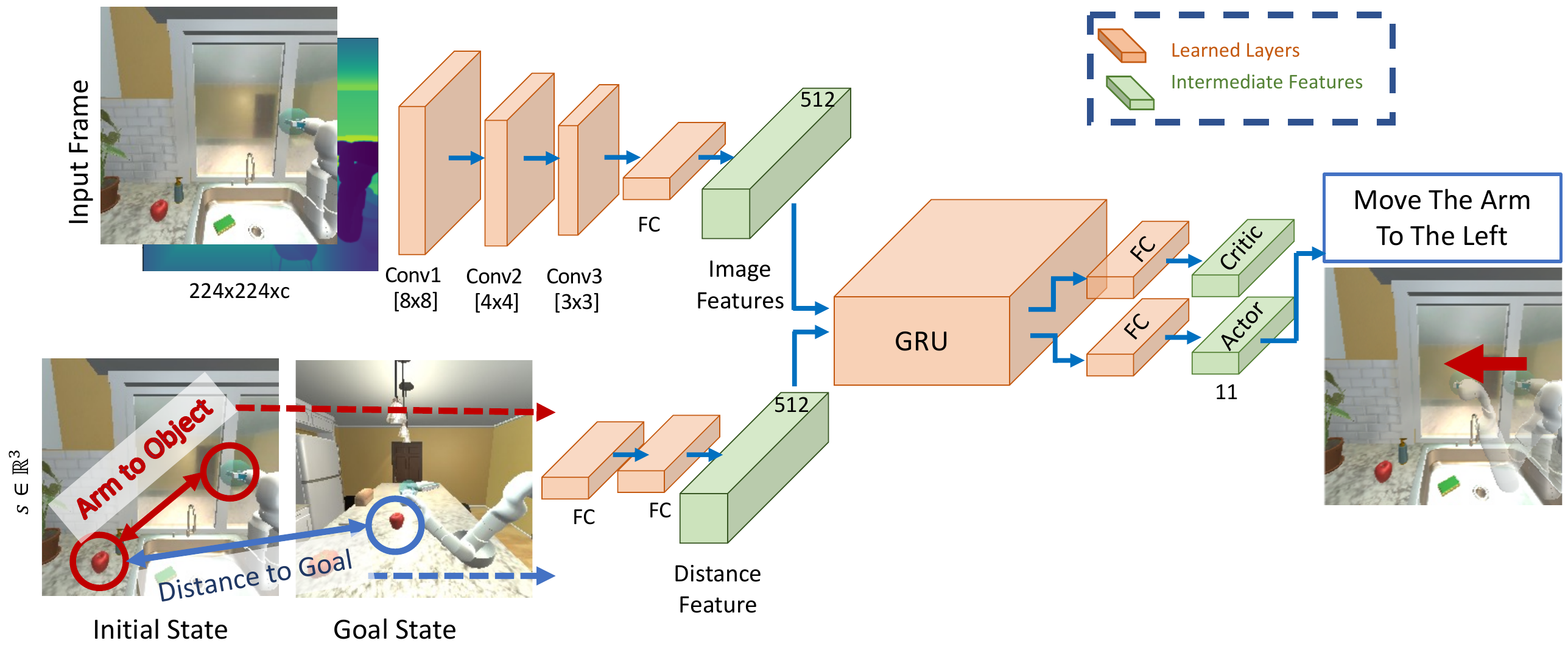}
    \caption{\textbf{Architecture}. Our network uses the Depth (c=1), RGB (c=3) or RGBD (c=4) observations and the agent's arm relative location to the object to estimate the movements needed to pickup the target object and take it to the goal state while avoiding the unwanted collisions.}
    \vspace{-1em}
    \label{fig:model}
    
\end{figure*}

\task requires agents to learn to navigate themselves along the 2D floor while also learning to navigate their arm and objects within the 3D space around them. Past works for visual object manipulation tend to use modular designs to solve this problem~\cite{murali20206,srivastava2014combined,garrett2020pddlstream} -- for instance, employing object detection models, instance segmentation models, point cloud estimators, etc. and then feeding these outputs into a planner. In contrast, recent developments in the Embodied AI domain~\cite{Zhu2017VisualSP,ALFRED20} have demonstrated the benefits of using end-to-end learning-based approaches. In this work, we investigate end-to-end learning approaches for our task. See Sec.~\ref{sec:experiments} for results obtained by a single end-to-end model in comparison to a disjoint approach.

Our approach builds upon the model and learning methods used in \cite{wijmans2019dd} for the task of \pnav. \armthor provides the agent with access to a variety of sensors including egocentric RGB and Depth sensors, GPS and Compass coordinates of the arm and target locations in the scene, ground truth instance segmentation maps as well as the kinematics of the arm. In this work, we investigate using the Depth and RGB sensors as well as the GPS and Compass coordinates, but leave other sensing modalities to future work.

Our agent at each time step uses as inputs, its egocentric frame $I_t$, the current relative distance of the arm (the end-effector location) to the object's location $d^o_{arm}$ and the current relative distance of the arm to the object's goal state $d^{goal}_{o}$. These observations are encoded into visual and location embeddings respectively and then fed into a controller to produce a distribution over possible actions, i.e. its policy. The discretized action space of the agent includes: moving forward, rotating the agent (left and right), pickup up the object, issuing done action, moving the arm in the space in front of the agent (ahead, backward, right, left, up, down), rotating the arm (in terms of Euler angles) and increasing or decreasing the height of the arm. Discretization of the action space is discussed in details in Sec.~\ref{sec:experiments}. An action is sampled from this policy and fed into the \armthor simulator which generates the next state and corresponding observations. An episode terminates when the object reaches its goal state or when the agent runs out of time (the episode reaches its maximum length).

Figure~\ref{fig:model} provides an overview of the model architecture. Visual embeddings are obtained using 3 convolution layers followed by a fully-connected layer with non-linearity in between to produce a $512$ dimensional vector. The relative coordinates (arm $\rightarrow$ object and arm $\rightarrow$ goal) are embedded by 2 fully connected layers with ReLU non-linearity to a feature vector of size $512$, which is then concatenated with the image features. The controller consisting of a GRU~\cite{cho2014learning} with a hidden size of $512$ uses this resulting embedding as input to produce a policy (i.e. distribution over actions) and a value (i.e. an estimate of future rewards).

The network parameters are optimized using DD-PPO~\cite{wijmans2019dd} with both a terminal reward and intermediate rewards used to shape the reward space. More specifically, the reward at each time step t is: 
\begin{equation}
    r_t = R_{success}.\mathbb{I}_{success} + R_{pickup}.\mathbb{I}_{pickup} + \Delta_{arm}^o + \Delta_{o}^{goal},
\end{equation}
where $R_{success}=10$, $R_{pickup}=5$, $\mathbb{I}_{success}, \mathbb{I}_{pickup}$ are the indicators of the success of the task and success of the object pickup, respectively, and $\Delta_{arm}^o$ and $\Delta_{o}^{goal}$ are the differences in the distance of the arm to object ($d^o_{arm}$), and the distance of the object to the goal ($d^{goal}_{o}$) in comparison to the previous timestep.
This method of reward shaping provides us the ability to balance the importance of the different phases of the task -- pickup vs place.

%% file: text/06_experiments.tex
\section{Experiments}
\label{sec:experiments}

\begin{figure*}[tp]
    \centering
    \includegraphics[width=\textwidth]{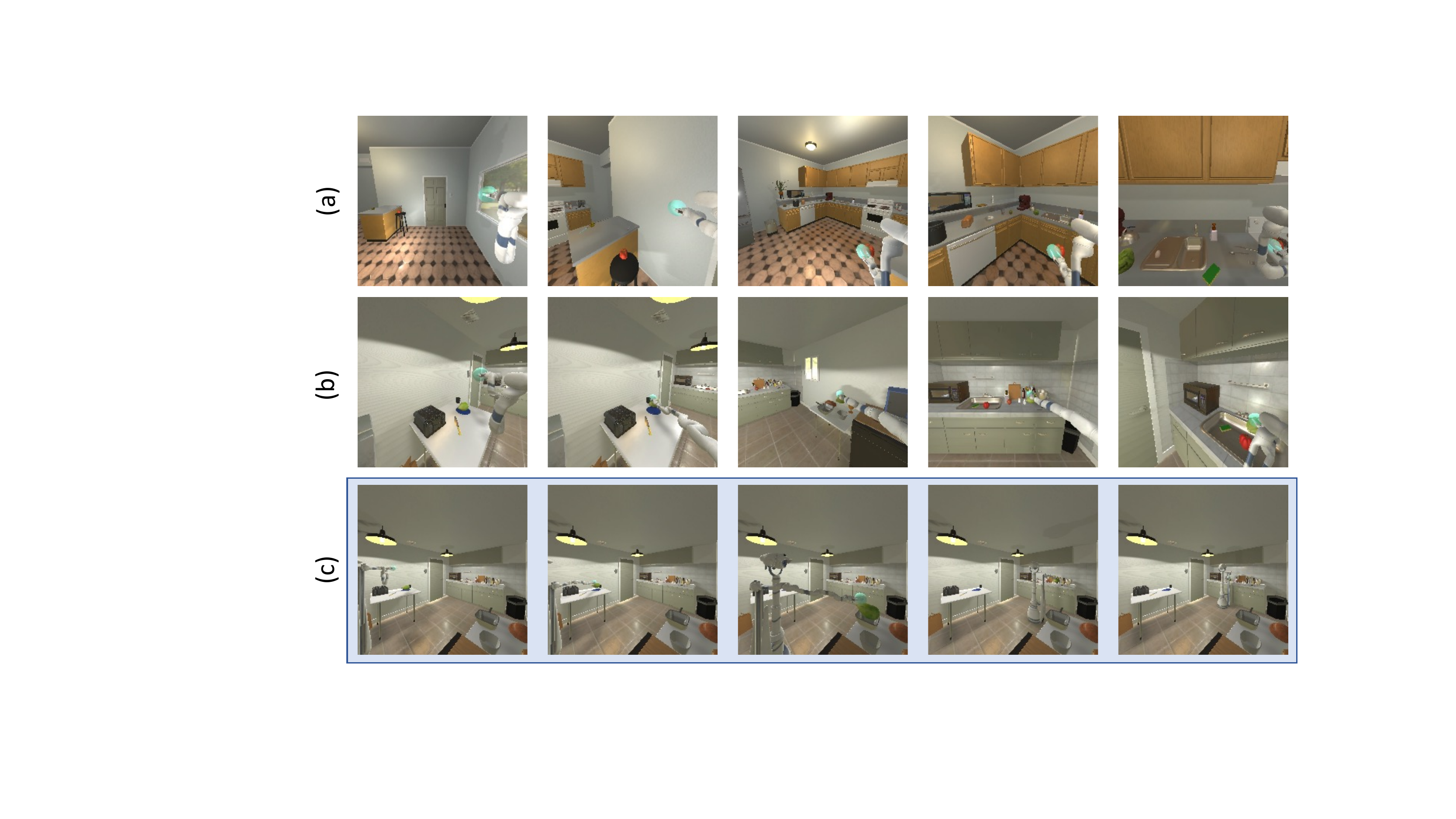}
    \caption{\textbf{Qualitative Results.} Our qualitative results illustrate our network's ability to generalize to picking up objects in novel environments and navigating the agent with the object in hand to the desired location. For example, row (c) shows a scenario from a third-perspective in which the agent needs to navigate in the kitchen while avoiding the collision with other objects (toaster, tomato and sink in this case).
    }
    \label{fig:qualitative}
    \vspace{-1em}
\end{figure*}

We now present results for our \task using the \dataset dataset and quantify its ability to generalize to new scenes as well as new objects within these scenes. We also provide comparisons of our end-to-end approach to a multi-stage model. Finally, we ablate the importance of the RGB and Depth sensors in the presence of the GPS and Compass coordinates. 

\noindent{\textbf{Experimental Setup.}} We use the AllenAct ~\cite{weihs2020allenact} framework to implement and train our models. Our agent uses an action space with $13$ discrete actions: 1) moving the agent forward by $20cm$, 2-3) rotating the agent to the right/left for $45$ degrees, 4-9) changing the relative location of the wrist by $5cm$ w.r.t. agent ($\pm x, \pm y, \pm z$), and 10-11) increasing or decreasing the height of the arm by $7cm$ (refer to Figure~\ref{fig:arm_constraint} for illustrations of the possible arm movements), 12) abstract grasp (which can be either successful or unsuccessful), 13) issuing a \textit{Done} action indicating the end of an episode. We use DD-PPO~\cite{wijmans2019dd} as our optimization algorithm. We train each model for 20M frames and maximum episode length of $200$.

As specified in Section~\ref{sec:dataset} we use 20 scenes for training, 5 for validation and 5 for test. Unless otherwise specified, we train our networks on the training scenes with 6 objects and evaluate the learned policy on the val and test scenes using the 6 seen and 6 novel object categories. Each object category includes instances of different shape and appearance.

\noindent\textbf{Quantitative results.} 
Table~\ref{tab:normal_method} reports results for our model using 6 metrics --  Episode Success w/o Disturbance (SRwD), PickUp Success (PuSR), Episode Success (SR), Ep-Len for PickUp (PuLen), Ep-Len for Success (SuLen) and Episode Length (Len). The proposed model performs reasonably well and achieves 39.4\% SRwD (68.7\% when allowing disturbance) for objects it has interacted with at training time. It obtains a significantly higher PuSR of 89.9\% indicating that navigating towards the initial location of the object and picking it up is easier than navigating the object through the scene. Interestingly, the model also generalizes moderately well to novel objects (row 2 -- Test-NovelObj). The SRwD metric degrades in the challenging zero shot setting compared to the seen object scenario. These results are promising and a stepping stone towards learning a generalizable object manipulation model. We also provide results for SeenScene-NovelObj -- where the agent is tested within seen environments but provided with novel objects. It is interesting to note that the performance is similar to Test-NovelObj, once again showing the challenges of moving to new object categories. Moving and manipulating objects requires an understanding of the object geometry (to be able to avoid collisions), and therefore generalizing to unseen objects is challenging.

\begin{figure}[tp]
    \centering
    \includegraphics[width=20pc]{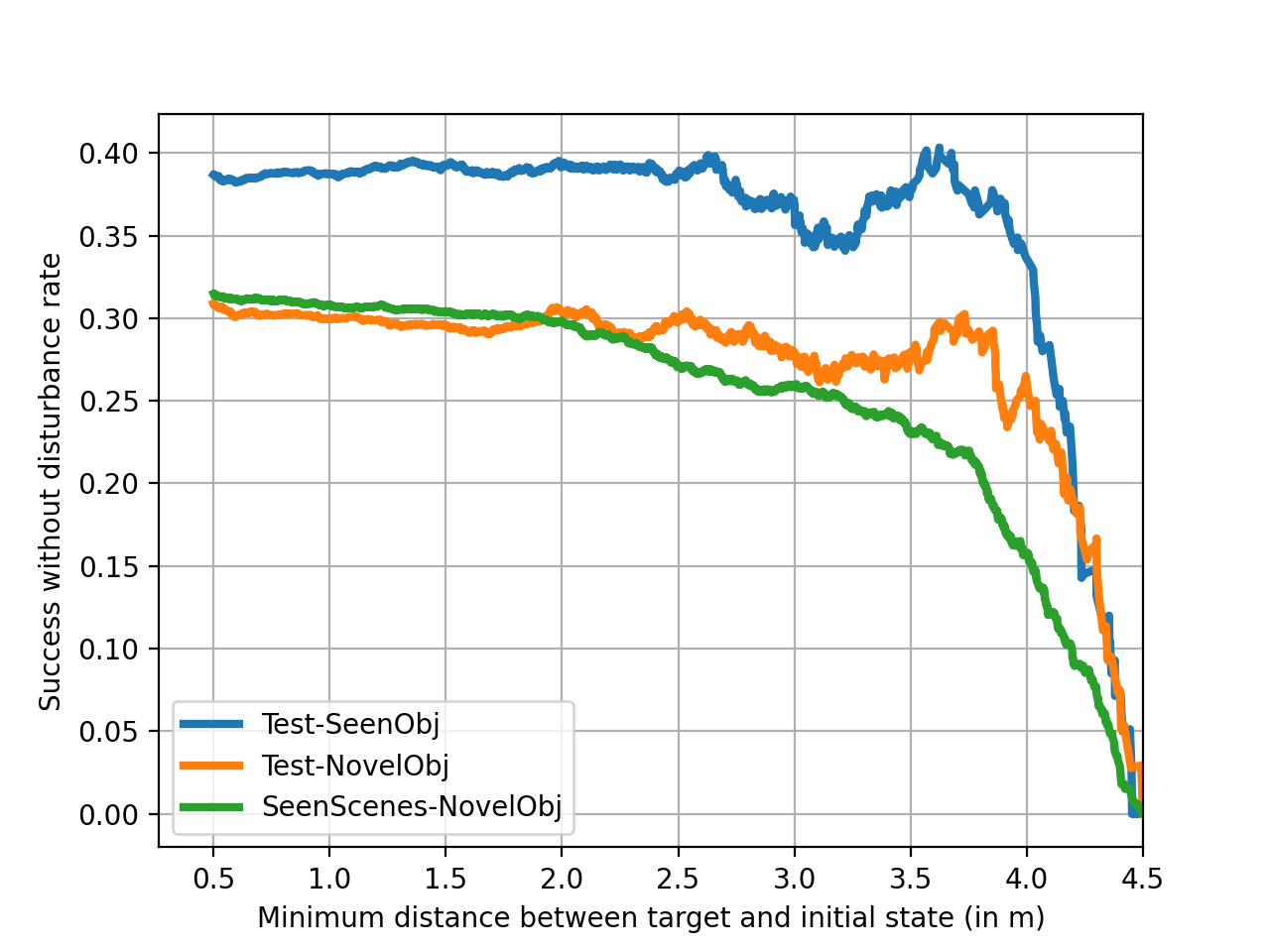}
    \caption{\textbf{Comparison of SRwD rates based on the initial distance to goal.} 
    The plot shows that our network's generalizability to novel scenes is superior to generalization to unseen objects.
    }
    \label{fig:distance_on_success}
    \vspace{-2em}
\end{figure}

\begin{table*}[tp]
\footnotesize
\setlength{\tabcolsep}{4pt}
	\centering
\hfill
	\begin{tabular}{cccccccc}
    &Ep-Success w/o Disturbance\%&PickUp Success \%&Episode Success\% &Ep-Len for PickUp&Ep-Len for Success&Ep-Len\tabularnewline
    &(SRwD)&(PuSR) &(SR) &(PuLen)&(SuLen)&(Len)\tabularnewline
\toprule
Test-SeenObj &39.4&89.9&68.7&43.6&78.1&114.0\tabularnewline
\midrule
Test-NovelObj&32.7 &84.3&62.1&48.1&82.4&122.0\tabularnewline
\midrule
SeenScenes-NovelObj &32.2&90.6&74.6&44.6&80.7&104.0\tabularnewline
\bottomrule

	\end{tabular}
\hfill 
\vspace{-1em}
	\caption{\textbf{Quantitative results.} The performance of our network on different data splits. Our experiments show that our trained agent can generalize to novel scenes and objects.}
	\label{tab:normal_method}
	\vspace{-1em}
\end{table*}
It is interesting to analyze the performance of this model and see how the performance of our method changes based on the closeness of the target location to the initial state. As expected, the SRwD drops as we increase the distance to the goal, since it becomes harder to navigate to the target location. However, our model is relatively robust to longer distances (Figure~\ref{fig:distance_on_success}). But once again, we see that the SRwD is higher for seen objects than novel objects.

Our qualitative results in Figure~\ref{fig:qualitative} illustrate a few examples of our agent's behavior on Test-SeenObj. {In the second row the agent must move a moderately sized object (lettuce) while in the first row it must move a smaller object (apple). The second row shows an example where the target location is on a table that has several other objects on it, necessitating careful movements of the arm while placing the object down.} The third row shows the episode in row (b) from a third camera view. Note that this view is purely for illustrative purposes and not available to the agent.

\noindent\textbf{Comparison to a Disjoint Model.} 
In this work we use a single end-to-end model for the entire task which includes navigating the arm to the object to pickup, and navigating the agent, arm and the object to the goal location. We posit that navigating the arm to pick up an object shares a lot of underlying physical understanding with moving the object within the 3D environment and it would help the performance to share weights among them. To evaluate this hypothesis, we also train a disjoint model -- which consists of two separate visual encoders and controllers, one for each sub-task of picking up the object and moving towards the goal (weights are not shared). At training time, the first sub-task model starts receiving gradients at the beginning of training, but the second sub-task model only receives gradients once the training episodes proceed beyond successful pickups. The results in Table~\ref{tab:ablation} show that having a disjoint approach improves PuSR and PuLen since the first model is only required to perform the simpler task of picking up the object. However, this model fails to learn to navigate the object to the goal since the model only receives gradients for a fraction of the training episodes (the ones that completed a pickup). Since it shares no parameters between the two phases, it cannot leverage skills and abilities learned across the two phases. While the number of parameters has doubled for the disjoint model, the training has also become less efficient. We acknowledge that the success rate might increase if the model is trained for longer\footnote{Training for 40M frames (2x normal training) slightly increased the success rate on training set but no increase on test set}. This ablation justifies our design choice for combining the two subtasks of \textit{pickup} and \textit{move-to-target}.

\noindent\textbf{No-Vision Agent.} 
Visual information is an important aspect of \task, in spite of having access to GPS and compass sensors. To illustrate the importance of the visual information in our training, we trained a model with no visual modalities but still with other sensory information such as the arm relative locations. In Table~\ref{tab:ablation}, we show that our agent outperforms the non-visual baseline by more than 2x, improving SRwD from $10.3$ to $39.4$.

\noindent\textbf{RGB and Depth modalities.}
\armthor provides a host of sensor modalities to train and evaluate agents. Our experiments thus far have only leveraged ego-centric Depth observations. In Table~\ref{tab:ablation}, we show results for agents that are trained using RGB and RGBD frames as well.
For our RGB experiment, we use the same architecture as Depth setup and for the RGBD input, we concatenate image features with depth features and use the combined feature as the input to the GRU. The rest of the setup is similar to our baseline method.
We observe that the depth only model outperforms the RGB model. A similar trend has been shown before for the PointNav task \cite{savva2019habitat}.
More complex networks and/or training paradigms might help improve metrics, and we leave this direction for future work. 

\begin{table}[tp]
\footnotesize
\setlength{\tabcolsep}{4pt}
	\centering
\hfill
	\begin{tabular}{cccccccc}
   &Sensors&SRwD &PuSR &SR &PuLen&SuLen&Len\tabularnewline
\toprule
No-Vision&G &10.3&66.8&18.7&38.1&65.6&64.9\tabularnewline
\midrule
Disjoint  &GD&0.0 &\textbf{91.6}&0.0&40.8&-&5.05\tabularnewline
Model&&&&&&\tabularnewline
\midrule
RGB&GR &21.2&68.3&37.7&56.2&91.1&53.0\tabularnewline
\midrule
RGBD&GRD&37.1 &86.8&62.8&45.1&82.2&123.0\tabularnewline
\midrule
Depth&GD&\textbf{39.4} &{89.9}&\textbf{68.7}&43.6&78.1&114.0\tabularnewline
\bottomrule

	\end{tabular}
\hfill 
\vspace{-1em}
	\caption{\textbf{Ablation Studies.} We study ablations of our network using different combinations of sensors as well as architecture design. Input sensors that are used by these networks are a subset of GPS (G), Depth (D) and RGB (R) sensors.}
	\label{tab:ablation}
	\vspace{-2em}
\end{table}

%% file: text/07_discussion.tex
\section{Discussion / Conclusion}
We propose \armthor a framework for visual object manipulation that builds upon the AI2-THOR environment. \armthor provides a host of diverse and visually complex scenes with several different object categories to manipulate. We hope that this new framework encourages the Embodied AI community towards tackling exciting problems dealing with object manipulation.
Using \armthor, we study the problem of \task, where the goal is to pick up an object and move it to a target location. Our experimental evaluations show that the state-of-the-art models that perform well on embodied tasks such as navigation are not as effective for object manipulation, indicating there is still a large room for improving models for this challenging task. 
Furthermore, in this paper we use GPS and visual sensors. Relaxing the usage of these sensors is an interesting task that is left for future work.

%% file: text/08_supplementary.tex
\clearpage
\appendix
\section*{Appendix}

\section{Data generation}
\label{sec:data_collection}

Our data generation process starts by finding all the possible locations in each room per object. Furthermore, we need to separate out the initial object locations for which there exists a valid agent pose that can reach it. This is to ensure the objects in each task of our dataset are reachable by our agent, and there is at least a solution for picking up/ dropping off the object in that location. Note that the object in the aforementioned target location can be visible or hidden from the agent's point of view (e.g., on a table vs. in the closed fridge). In this paper we only consider the problem for visible objects. Since there is a path between any two agent locations in the room, any pair of possible locations can be considered as a task (which may or may not require navigating the agent in the scene).

\section{More Dataset Stats}

\begin{figure}
    \centering
    \includegraphics[width=15pc]{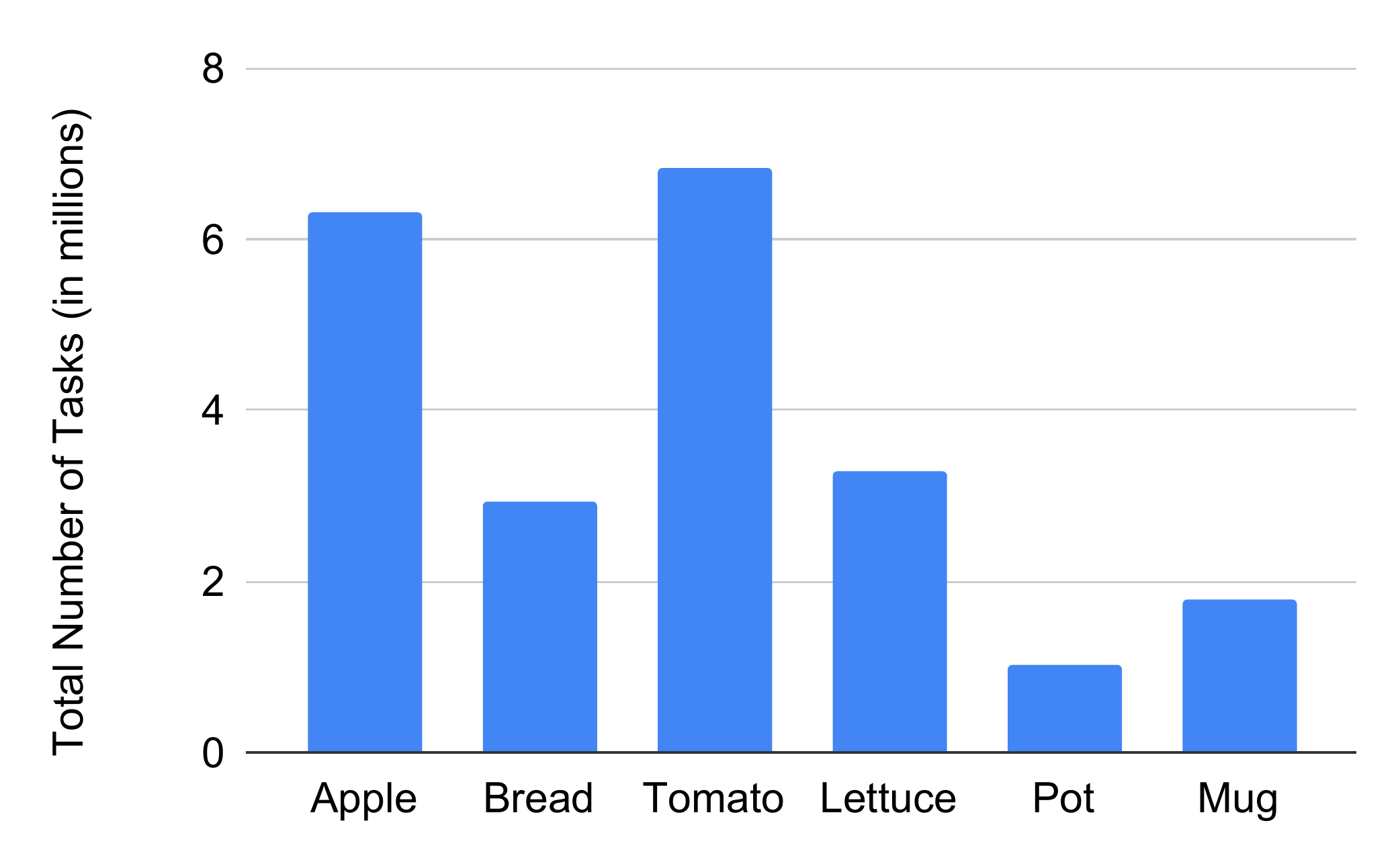}
    \vspace{-1em}
    \caption{\textbf{Number of possible tasks.}
    }
    \vspace{-1em}
    \label{fig:dataset_stat_tasks}
    
\end{figure}

In Section 4, we discussed details and some statistical aspects of the dataset. In Figure 3 in the main paper, the training data split is described as as a large streaming pool of tasks that can be sampled from. In Figure~\ref{fig:dataset_stat_tasks}, we provide the number of tasks (a pair of possible object locations described in the previous section) per training object. Due to the physical space constraints, the number of possible object locations (and as a result the number of possible tasks) for smaller objects is higher than the bigger ones (e.g., Apple vs. Pot).

In Table~\ref{tab:split_object_stats}, we illustrate the number of possible object locations per object for each data split. This emphasizes on the diversity of our dataset.

\begin{table}
\footnotesize
\setlength{\tabcolsep}{4pt}
	\centering
\hfill
	\begin{tabular}{ccccc}
    &Train&Test&Val&Total\tabularnewline
	    \toprule
Apple&12523&2858&2012&17393\tabularnewline
	    \midrule
Bread&8003&1612&1817&11432\tabularnewline
	    \midrule
Tomato&12158&2641&2424&17223\tabularnewline
	    \midrule
Lettuce&7942&1267&1454&10663\tabularnewline
	    \midrule
Pot&4298&918&937&6153\tabularnewline
	    \midrule
Mug&6289&1555&1586&9430\tabularnewline
	    \bottomrule
	\end{tabular}
\hfill 
\vspace{2mm}
\vspace{-1em}
	\caption{\textbf{Possible location of objects.} The distribution of possible object locations per object per data split. In total, there are 72k object locations in our dataset.}
	\vspace{-1em}
	\label{tab:split_object_stats}
\end{table}

\begin{figure*}[tp]
\centering
\includegraphics[width=40pc]{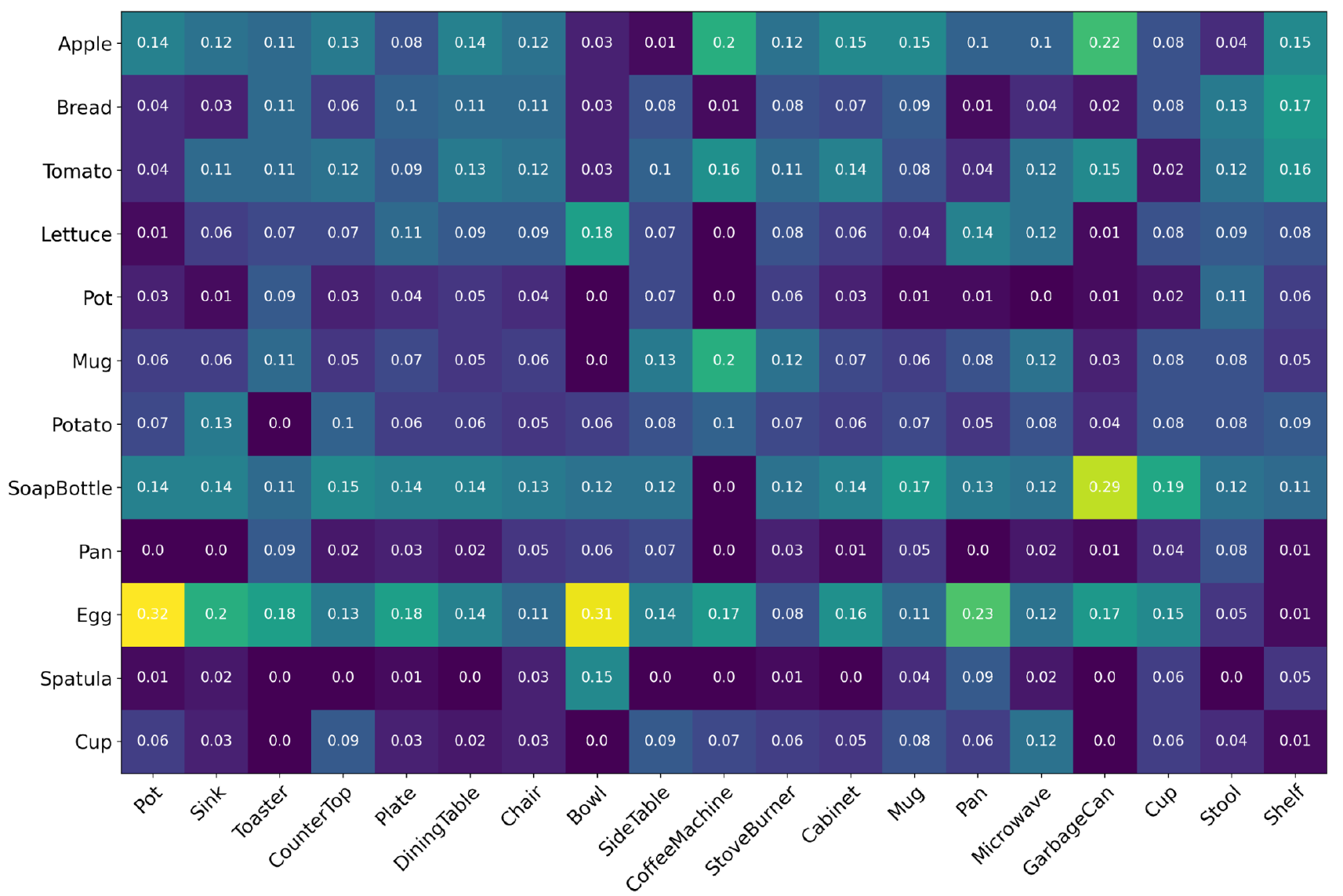} 
\vspace{-1em}
\caption{\textbf{Co-occurrences of objects and receptacles.} The heatmap provides the distribution of object locations on each receptacles and illustrates the diversity of our dataset.}
\vspace{-1em}
\label{fig:dataset_coocur}
\end{figure*}

Figure~\ref{fig:dataset_coocur} shows the co-occurrences of objects and their source and target receptacles. As seen, objects may lie in a diverse set of locations, requiring the agent to be able to reach for and place objects in a variety of situations.

